\documentclass[sigconf,table]{acmart}
\usepackage{graphicx}
\usepackage{rotating}
\usepackage{xcolor}
\usepackage{hyperref}
\usepackage{wrapfig}
\AtBeginDocument{%
  }

\setcopyright{acmlicensed}
\copyrightyear{2025}
\acmYear{2025}
\acmDOI{}
\acmConference[TACPS 2025]{Proceedings of 3rd International Workshop on Trustworthy Autonomous Cyber-Physical Systems}{July 21, 2025}{Zagreb, Croatia}
\acmISBN{}

\begin{document}

\title{Neurosymbolic Feature Extraction for Identifying Forced Labor in Supply Chains}

\author{Zili Wang}
\orcid{0000-0003-1730-6180}
\affiliation{%
  \institution{Iowa State University}
  \department{Department of Computer Science}
  \city{Ames}
  \state{Iowa}
  \country{USA}
}
\email{ziliw1@iastate.edu}
\author{Frank Montabon}
\affiliation{%
  \institution{Iowa State University}
  \department{Department of Supply Chain Management}
  \city{Ames}
  \state{Iowa}
  \country{USA}
}
\email{montabon@iastate.edu}
\orcid{0000-0003-3561-7433}
\author{Kristin Yvonne Rozier}
\orcid{0000-0002-6718-2828}
\affiliation{%
  \institution{Iowa State University}
  \department{Department of Aerospace Engineering}
  \city{Ames}
  \state{Iowa}
  \country{USA}
}
\email{kyrozier@iastate.edu}

\renewcommand{\shortauthors}{Wang et al.}

\begin{abstract}
  Supply chain networks are complex systems that are challenging to analyze; this problem is exacerbated when there are illicit activities involved in the supply chain, such as counterfeit parts, forced labor, or human trafficking.  
  While machine learning (ML) can find patterns in complex systems like supply chains, traditional ML techniques require large training data sets. 
  However, illicit supply chains are characterized by very sparse data, and the data that is available is often (purposely) corrupted or unreliable in order to hide the nature of the activities. 
  We need to be able to automatically detect new patterns that correlate with such illegal activity over complex, even temporal data, without requiring large training data sets. 
  We explore neurosymbolic methods for identifying instances of illicit activity in supply chains and compare the effectiveness of manual and automated feature extraction from news articles accurately describing illicit activities uncovered by authorities. 
  We propose a question tree approach for querying a large language model (LLM) to identify and quantify the relevance of articles. 
  This enables a systematic evaluation of the differences between human and machine classification of news articles related to forced labor in supply chains.
\end{abstract}




\maketitle

\section{Introduction}
\label{sec:intro}

Illicit supply networks facilitate the operation of major crime syndicates; their complexity and overlap with commercial supply chain communications, logistics, transportation, and financial infrastructures make illicit supply networks particularly hard to track down. 
Detection is difficult due to several factors, including the global nature of supply chains, profit-driven incentives, sparse or corrupted datasets, and a lack of transparency, necessitating the use of indirect data \cite{Singhal_2011}.  
For example, a consumer product may be assembled in one country, while its raw materials are sourced from another, and its labor provided by workers from a third.
Even if a company values ethical labor practices, it may be difficult to ensure that every supplier throughout the chain follows the same standards, and any practices that disrupt supply chain operation can be costly \cite{Hendricks_Singhal_2008}.

Recent research programs and policy initiatives have recognized this challenge and emphasized the need for automated, intelligent methods to detect illicit supply chains. 
The DARPA A3ML program calls for the development of algorithms that can extract illicit financial flow tactics, techniques, and procedures, with the goal of making it prohibitively expensive to launder value through the global financial system \cite{DARPA_A3ML_2025}.
Similarly, the SECUBIC project focuses on securing software supply chains at the binary level, addressing trust and integrity challenges in components by developing techniques for analyzing and verifying binaries throughout their lifecycle \cite{SECUBIC_2025}. 
The U.~S.~National Science Foundation called for cross-disciplinary teams to improve understanding of the operations of illicit supply networks and strengthen the ability to detect, disrupt, and dismantle them \cite{D-ISN}.
In addition, recent calls for research in formal methods have emphasized the need for verifiable and trustworthy AI systems in domains such as sustainable and safe supply chains, as illustrated by a 2025 FMNET announcement of a project focused on formal verification in agri-food systems \cite{MuFMNETEmail2025}.

Efforts to prevent forced labor have also led to legislation, such as Iowa Code 80.45A (human trafficking prevention in lodging providers) \cite{Iowa-code}, the U.K. Modern Slavery Act of 2015, and the California Transparency in Supply Chains Act of 2010, which aim to improve corporate accountability and transparency \cite{Gold2015,New2015,Stevenson2018}. 
However, enforcement of these policies remains limited in practice, due in part to the challenge of creating accurate, scalable detection methods.
The problem is further complicated by the fluid nature of worker exploitation, where individuals may transition between different forms of abuse and coercion within a short period of time \cite{LeBaron2021}, or be hidden in plain site, even at universities \cite{Red21,Hytrek_2021}.

To address this gap, recent work has explored the use of automated text analysis to identify indicators of forced labor and related illicit activities.
For example, \cite{RaFoLa} contributes a corpus of 989 news articles, annotated by both human experts and machine intelligence, to identify risk factors based on International Labour Organization standards.
This work highlights the potential of explainable multi-class classification in surfacing indicators of forced and slave labor from open-source text.
In practice, having clear, interpretable justifications for classification is essential from both legal and ethical perspectives.

Large language models (LLMs) have successfully detected other forms of illicit activity.
For example, LLMs identified drug trafficking activity on social media, using prompting frameworks designed to handle deceptive language and euphemisms \cite{Hu2024}.
Similarly, \cite{Pendyala2024} investigates how LLMs can be used to detect misinformation, which is relevant when analyzing the narratives surrounding labor practices and supply chain disclosures.
These examples show how AI systems can help identify hidden or obfuscated signals of exploitation, providing a foundation for similar techniques to be applied to forced labor detection.

We focus first on the problem of detecting forced labor in supply chains. 
In order to identify factors of forced labor in supply chains, we propose manual and automated methods for feature extraction.
We seek to compare the effectiveness of these methods in identifying relevant articles, as well as combine them to improve the overall detection process.
To deepen the analysis, we explore a possible formal methods-based approach using Boolean formula enumeration to model complex relationships between the extracted features.
Our work lays the groundwork for a neurosymbolic framework that integrates large language models and formal reasoning to support scalable, interpretable detection of forced labor in supply chains.

\section{News Article Data Extraction Methods}
\label{sec:method}

We aim to compare and contrast how effective manual and automated methods are at identifying factors of forced labor in supply networks, as well as how these methods can be effectively combined for reliable data aggregation.
We previously identified a set of $25$ features that could be indicators of forced labor instances \cite{Schumm2025}, which we detail in Table~\ref{tab:features}.
Our manual and automated methods extract these $25$ features from real instances of forced labor data, which we describe in Sections \ref{sec:manual} and \ref{sec:automated} respectively.
Although these features are not individually condemning evidence of forced labor practices, more complex patterns and relationships between these variables may indicate forced labor in supply chains.
To this end, we detail how we can apply a SAT-based Boolean formula enumeration technique from the literature \cite{Schumm2025} to analyze the relationships between the extracted features in Section \ref{sec:boolean}.

\subsection{Manual Feature Data Extraction}
\label{sec:manual}

To manually identify a candidate set of news articles related to forced labor in supply chains, we query the online news databases ProQuest \cite{lexisnexis} and LexisNexis \cite{proquest} using the search terms ``forced labor'' and ``supply chain,'' gathering over $340$ articles from between the years $2016$ to $2024$.
We then classify these articles by hand as being relevant or irrelevant to forced labor in supply chains, and extract from the relevant articles the $25$ features that we previously identified as being indicative of forced labor in supply chains.
This resulted in a total of $125$ \textit{incidents} that were manually classified as relevant instances of forced labor in supply chains.
We define an incident as a single instance of forced labor in a supply chain, which may be reported in multiple articles.

Our final dataset aggregates incidents from a large variety of supply chain industries including textiles, seafood and agriculture, precious metals, sexual services, and steel. 
For instance, one incident \cite{ejf2025nk} involves a fleet of Chinese tuna fishing vessels reportedly using North Korean forced laborers to illegally fish in the Indian Ocean.
For this incident, we label the associated product as ``tuna,'' the supply chain as ``seafood,'' the industry as ``commercial fishing,'' the country of incident as ``China,'' and marked the Boolean features of High Risk Sourcing Country and High Risk Product as \textit{True}.
In another incident \cite{usdoj2021bloomingonion}, twenty-four individuals were indicted for forcing trafficked Mexican and Central American workers to perform grueling physical work picking onions, and housing them in crowded, unsanitary conditions.
For this incident, we label the associated product as ``onions,'' the supply chain as ``Cash Crops/Produce,'' the industry as ``agriculture,'' and marked the Boolean features of High Risk Sourcing Country and High Risk Product as \textit{False}.

\begin{table*}
  \centering
  {\fontsize{8.5}{9}\selectfont
  \setlength{\tabcolsep}{3pt}
  \renewcommand{\arraystretch}{1.1}
  \begin{tabular}{lll}
    \rowcolor[HTML]{D9D9D9}
    \textbf{Variable} &
      \textbf{Type of Data} &
      \textbf{Definition / Explanation} \\
    Company &
      String &
      Only if a specific firm is portrayed as being at fault. \\
    \rowcolor[HTML]{D9D9D9} 
    Product &
      String &
      The product associated with this supply chain incident. \\
    Supply Chain &
      String &
      The primary end product or service the supply chain produces. \\
    \rowcolor[HTML]{D9D9D9} 
    Date of Incident &
      Date range &
      Reported date range of incident. \\
    Industry &
      String &
      \begin{tabular}[c]{@{}l@{}}The primary industry of the firm involved in forced labor, \\ which may be different from "Supply Chain" variable.\end{tabular} \\
    \rowcolor[HTML]{D9D9D9} 
    SIC &
      String &
      Standard Industrial Classification code. \\
    Sourcing Characteristic &
      Categorical &
      \begin{tabular}[c]{@{}l@{}}One of: Agriculture, Fishing, Mining, Sex Work, \\ Clothing/Textiles, or Hospitality.\end{tabular} \\
    \rowcolor[HTML]{D9D9D9} 
    \begin{tabular}[c]{@{}l@{}}Product or service \\ crosses national \\ border\end{tabular} &
      Boolean &
      \begin{tabular}[c]{@{}l@{}}Article must make clear that product or service crosses \\ a national border.\end{tabular} \\
    Age of Company &
      Int & Number of years since the company was founded.
       \\
    \rowcolor[HTML]{D9D9D9} 
    Country of Incident &
      String &
       \\
    \begin{tabular}[c]{@{}l@{}}State/Region/Province \\ of Incident\end{tabular} &
      String &
       \\
    \rowcolor[HTML]{D9D9D9} 
    \begin{tabular}[c]{@{}l@{}}High Risk \\ Sourcing Country\end{tabular} &
      Boolean &
      \begin{tabular}[c]{@{}l@{}}Corruption index between 0 and 50 of country where incident \\ happened (see https://www.transparency.org/en/cpi/2023).\end{tabular} \\
    High Risk Product &
      Boolean &
      \begin{tabular}[c]{@{}l@{}}If it appears on this (List of Goods Produced by Child Labor \\ or Forced Labor | U.S. Department of Labor (dol.gov)) \\ spreadsheet, then "Y."  If U.S., "N/A."\end{tabular} \\
    \rowcolor[HTML]{D9D9D9} 
    \begin{tabular}[c]{@{}l@{}}Concerns or Evidence \\ of Fake/Forged \\ Documentation\end{tabular} &
      Boolean &
       \\
    Position in Supply Chain &
      Int &
      \begin{tabular}[c]{@{}l@{}}1 = Extraction.\\  2 = Supplier (value-added after extraction).\\  3 = Manufacturer (product in its final form is created).\\  4 = Distribution of completed product.\end{tabular} \\
    \rowcolor[HTML]{D9D9D9} 
    Raw Material Supplier &
      Boolean &
      Does the company supply raw materials? \\
    Firm provided housing &
      Boolean &
      Does the company provide housing for its employees? \\
    \rowcolor[HTML]{D9D9D9} 
    Firm provided transportation &
      Boolean &
      Does the company provide transportation for its employees? \\
    Forced Labor Detected &
      Boolean &
      \begin{tabular}[c]{@{}l@{}}Forced labor is when workers are unable to control their \\ employment situation, whether through coercion or lack of \\ other options, with the worker effectively having no other \\ means of employment choice.\end{tabular} \\
    \rowcolor[HTML]{D9D9D9} 
    Slave Labor Detected &
      Boolean &
      \begin{tabular}[c]{@{}l@{}}Situations where person can't opt out, ownership of person is\\ implied or explicit, or location of residence and work is\\ dictated by someone else.\end{tabular} \\
    Child Labor Detected &
      Boolean &
      \begin{tabular}[c]{@{}l@{}}Person is considered a child (minor) by law of the country \\ where labor is taking place and law does not have exception \\ for such labor (e.g., farm families) or labor meets qualities \\ listed at https://www.ilo.org/topics-and-sectors/child-labour.\end{tabular} \\
    \rowcolor[HTML]{D9D9D9} 
    Mandatory Overtime &
      Boolean &
      \begin{tabular}[c]{@{}l@{}}Hourly employee can't opt out of overtime. \\ Overtime laws depend on country.\end{tabular} \\
    Sex Trafficking &
      Boolean &
      \begin{tabular}[c]{@{}l@{}}Person has not voluntary agreed to be a sex worker. \\ "Voluntary" means that no coercion is involved.\end{tabular} \\
    \rowcolor[HTML]{D9D9D9} 
    Prison Labor- Voluntary &
      Boolean &
      \begin{tabular}[c]{@{}l@{}}Person is being detained by a government regardless of reason. \\ Person has ability to opt in or out of labor.\end{tabular} \\
    Prison Labor - Forced &
      Boolean &
      \begin{tabular}[c]{@{}l@{}}Person is being detained by a government regardless of reason. \\ Person can't opt out of labor.\end{tabular}
  \end{tabular}
  }
  \caption{Data classification for 25 features that are possibly correlated with forced labor in supply chains.}
  \label{tab:features}
\end{table*}

\subsection{Automated Feature Data Extraction}
\label{sec:automated}

\begin{figure}
    \centering
    \includegraphics[width=\linewidth]{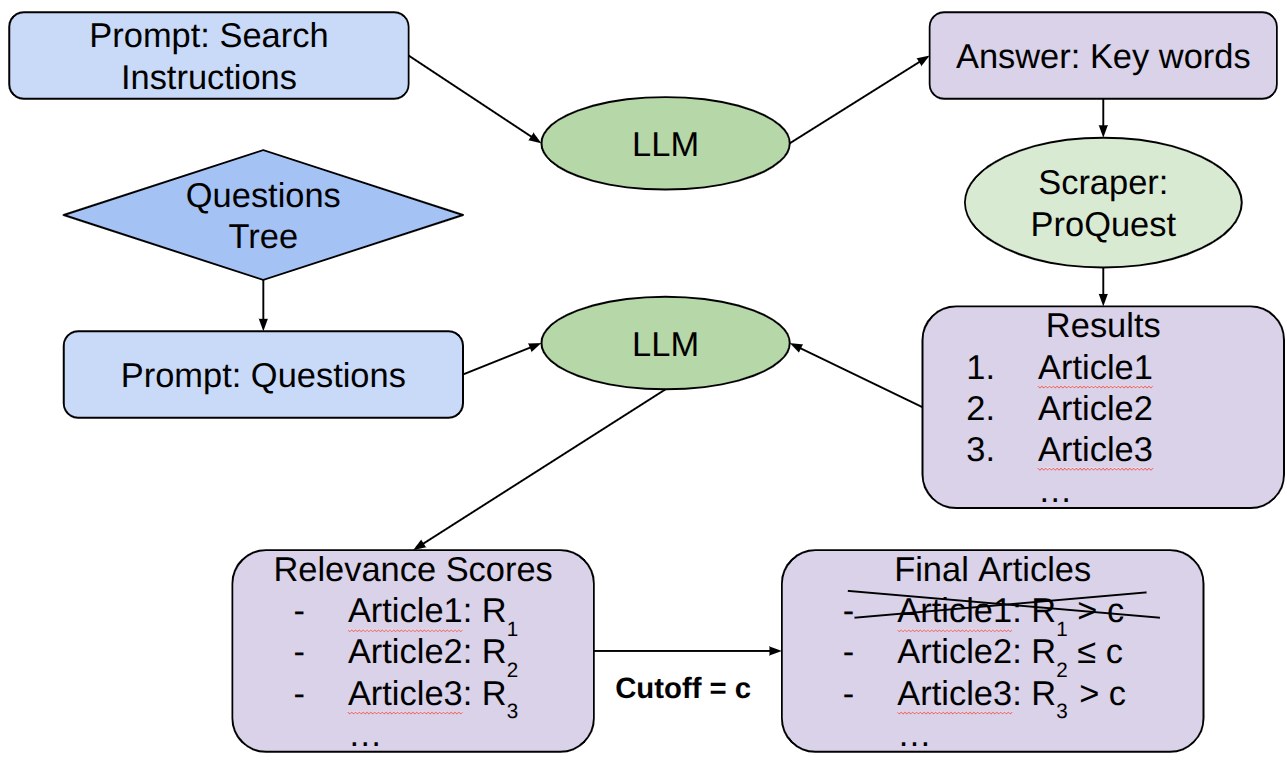}
    \Description{A diagram showing the automated methodology for identifying factors of forced labor in supply networks, including steps such as article search, web scraping, and relevance scoring.}
    \caption{
    The automated method involves searching for articles related to forced labor in supply chains, using the ProQuest database and a large language model (LLM) to query for relevant key words, scraping the web for articles, and then evaluating the relevance of each article to forced labor in supply chains using a question tree framework.
    }
    \label{fig:method}
\end{figure}

We prompt a large language model (LLM) to search for articles related to forced labor in supply chains by querying for relevant key words. 
We use the GPT-4.0 LLM, which is a state-of-the-art language model trained on a large corpus of text data \cite{openai2024gpt4technicalreport}.
We then use the ProQuest API
\footnote{We did not use the LexisNexis API in the automated method, as it is not freely available.}
to obtain a list of the articles matching these keywords.
Finally, we use a question tree framework \cite{qtree} to evaluate the relevance of each article to forced labor in supply chains, from which we calculate a relevance score for each article. We define a cutoff threshold to classify articles as relevant or irrelevant based on the relevance scores. These cutoff thresholds are defined based on the desired size of the dataset and the expected proportion of relevant articles.
Figure~\ref{fig:method} visualizes our automated feature extraction process.

\begin{figure*}[ht]
    \centering
    \includegraphics[width=0.8\textwidth]{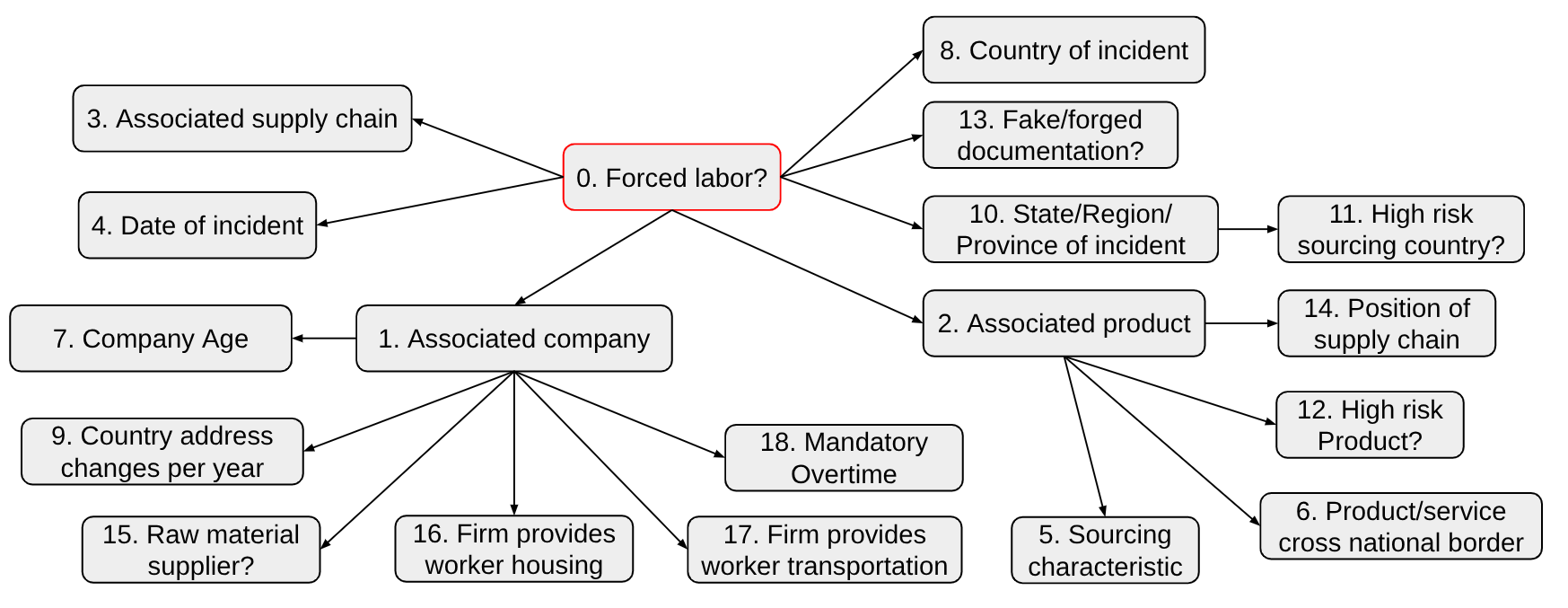}
    \caption{
    The question tree is a directed acyclic graph (DAG) where each node represents a question, and an edge from node $A$ to node $B$ indicates that a positive answer to question $A$ implies that question $B$ is relevant.
    For instance, we always start at the root node (outlined in red) of the question tree with the question ``Does the article mention forced labor?'' If the answer is yes, a score of $1$ is assigned to the root node, and then we proceed to evaluate the remaining questions in the children nodes.
    If the answer is no, the remaining questions are not evaluated, and a score of $0$ is assigned to each children node of the question tree.
    }
    \label{fig:qtree}
    \Description{
    A directed acyclic graph (DAG) representing a question tree, where each node represents a question, and an edge from node A to node B indicates that a positive answer to question A implies that question B is relevant. The root node is "Does the article mention forced labor?", and if the answer is yes, then we proceed to evaluate the remaining questions.
    }
\end{figure*}

\paragraph{Question Tree Framework.}
The question tree framework builds on the previously identified features of forced labor in supply chains, and uses a set of questions to evaluate the relevance of each article to forced labor.
We provide a visual representation of the question tree in Figure~\ref{fig:qtree}.
The question tree is a directed acyclic graph (DAG) where each node represents a question, and an edge from node $A$ to node $B$ indicates that a positive answer to question $A$ implies that question $B$ is relevant.
For instance, the root node of the question tree is the question ``Does the article mention forced labor?'' and if the answer is ``yes,'' then we proceed to evaluate the remaining questions. 
If the answer is ``no,'' the remaining questions are not evaluated, and a score of $0$ is assigned to each children node of the question tree.
As another example, the node for question $2$ checks that the article mentions a relevant good or product that is produced by forced labor; a negative answer to this question implies all the children nodes are irrelevant, as these questions further evaluate the relevance of the product to forced labor.

Using this framework, we can calculate a relevance score for each article by summing the scores of every node of the question tree, and dividing by the total number of nodes to normalize the score.
For features that we want to emphasize more, we may assign each node a weight, and normalize by the summed weights of all nodes instead.

\subsection{Boolean Formula Enumeration}
\label{sec:boolean}

\begin{figure}[htbp]
    \centering
    \includegraphics[width=0.81\linewidth]{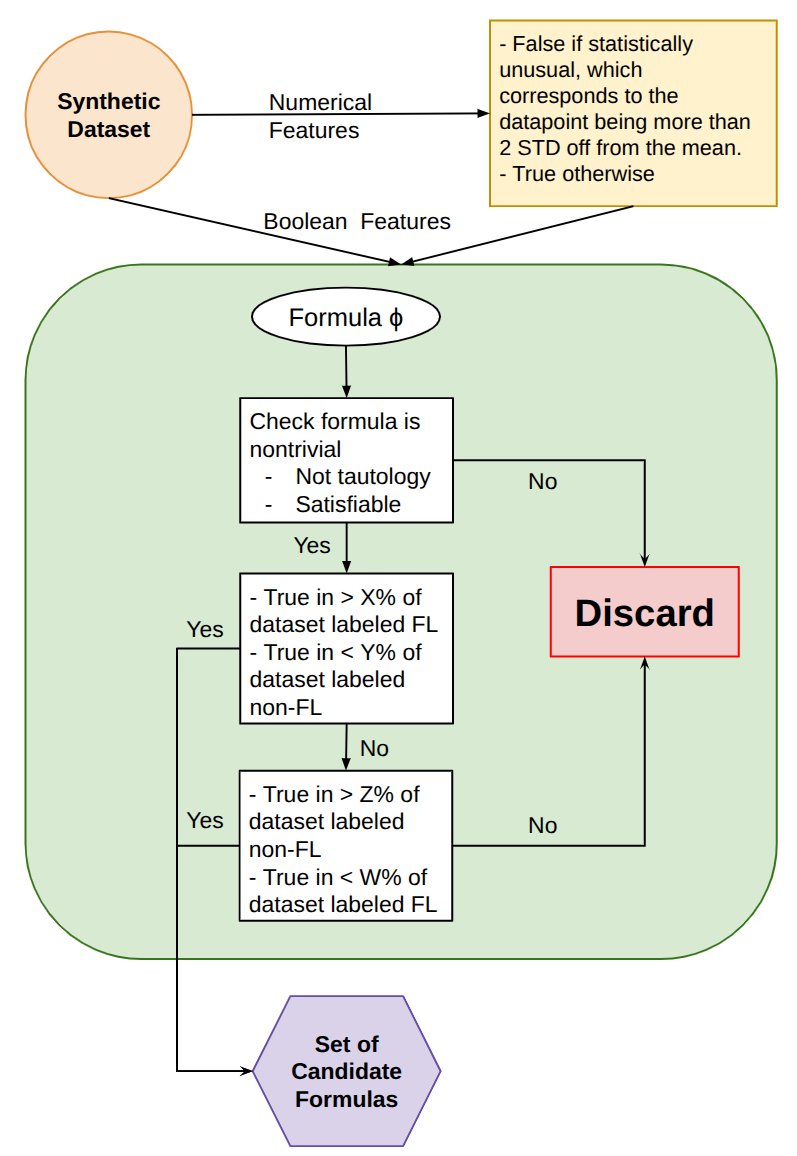}
    \caption{SAT formula enumeration technique used to identify the most relevant combinations of features for forced labor in supply chains on a synthetic dataset.
    }
    \label{fig:boolean_enum}
    \Description{
    A diagram showing the SAT formula enumeration technique used to identify the most relevant combinations of features for forced labor in supply chains on a synthetic dataset.
    }
\end{figure}

We can apply a SAT-based Boolean formula enumeration technique \cite{Schumm2025} to analyze the relationships between the extracted features.
Previously, we created a synthetic dataset of forced labor instances based on the $25$ features we identified in order to evaluate the effectiveness of this technique, which enumerates the most relevant combinations of features that are indicative of forced labor in supply chains.
First we encode the features as Boolean variables; numerical features are ``Booleanized'' by checking if they are more than two standard deviations away from the sample mean of that feature. 
We enumerate all possible Boolean formulas over this variable set, using a SAT solver to eliminate formulas that are not satisfiable or vacuously true, then compute how frequently each formula was satisfied by instances of forced labor in the synthetic dataset, and dissatisfied by non-instances of forced labor.
A formula that is satisfied by a large number of instances of forced labor, and dissatisfied by a small number of non-instances (or vice versa) is considered to be a meaningful formula that can be indicative of either the presence or absence of forced labor in supply chains.
We summarize this technique in Figure~\ref{fig:boolean_enum}. 
\emph{Note that while we started with purely Boolean formulas, this same algorithm can apply to temporal formulas.} 
For example, it may be valuable to check how rapidly or slowly certain features change as a product moves throughout a supply chain.

We apply this technique to the data collected from the manual and automated feature extraction methods to evaluate how realistic the previous synthetically generated dataset is, and to improve the fidelity of that synthetic dataset.
For instance, this technique applied to the previous synthetic dataset identified the following Boolean formula as being indicative of forced labor in supply chains:
$$\texttt{cross\_border } \land\ ( \texttt{high\_risk\_source } \lor \texttt{ high\_risk\_product})$$

Here, \texttt{cross\_border} indicates whether the product crosses a national border, \texttt{high\_risk\_source} indicates whether the sourcing country is considered high risk, and \texttt{high\_risk\_product}  indicates whether the product is considered high risk.
This formula indicates that the combined presence of either a high risk sourcing country or a high risk product, along with the product crossing a national border, can be indicative of forced labor in supply chains.

In order to apply this technique to the data collected from the manual and automated feature extraction methods, we would need non-instances of forced labor in supply chains to compare against.
Currently, the $125$ data-points we have collected are all instances of forced labor in supply chains, but this method needs to be able to compute how meaningful a formula is by the ratios of instances of forced labor that satisfy the formula, and non-instances of forced labor that dissatisfy the formula.
One possible solution is to use the same LLM that we used to query for articles, and use it to additionally classify and extract features from articles that were discarded as not being relevant instances of forced labor in supply chains.
This would allow us to generate a set of non-instances of forced labor in supply chains, that we can then apply the Boolean formula enumeration technique to.

\section{Future Research Directions}
\label{sec:conclusion}

Our vision for reliable illicit supply chain analysis addresses the challenge that sparse or unreliable data prevents using ML-based detection algorithms by extracting useful features from relevant news articles about supply chains, and involving formal methods in analyzing the relationships between these features.
Next steps involve expanding this work in several directions, including data collection, combining manual and automated methods, and applying other formal techniques to analyze the data.

\paragraph{Data Collection.}
We can improve data collection by adding more articles to the dataset, or by improving the quality of the articles. Querying for more keywords will result in more articles, but they may not be relevant, so we focus on improving the quality of the articles.
Having multiple people manually classify the same articles would help to improve confidence in the results, and also help to identify any inconsistencies in the manual feature extraction process.
We can apply the same philosophy to automated feature extraction, where we can take an ensemble of multiple LLMs to classify the same articles, achieving a similar effect. We may also explore other sources of indirect but reliable data, such as human experts or law enforcement data.

\paragraph{Combining Manual and Automated Feature Extraction.}
The manual and automated methods should be combined to provide a more effective solution to the problem.
For example, using the question tree framework to manually classify articles would help to ensure a more consistent classification compared to the automated method.
Such a process would also help inform what weights to assign to each node of the question tree, allowing us to leverage human domain-specific knowledge to improve the automated classification process.
One particularly relevant example of this is in how we iterated upon our previous work in \cite{Schumm2025}; one feature we previously identified was looking at the number of times that a company changed physical location.
Intuitively, a large number of changes in location may indicate that the company is trying to hide something.
However, this information was never found in news articles, and so we removed this feature from the question tree; effectively, the weight of this feature is $0$.

\paragraph{Formal Methods.}
We can apply Boolean enumeration techniques to analyze the data collected from the aggregation of the manual and automated methods.
We can examine the resulting Boolean relationships found between the relevant features, and try to use existing literature from the supply chain domain to explain the relationships.
It would be interesting to see if the relationships found in the synthetic dataset \cite{Schumm2025} are also present in the real data collected from the manual and automated methods. 

Furthermore, we can also expand the feature set to include temporal or epistemic features and then utilize temporal logic or epistemic logic formulas to check for relationships that correlate with illicit supply chains. 
For example, by comparing the time of employee recruitment with the time they begin work, we can detect unusually long delays, which may indicate that workers are being held in coercive conditions or transported across borders under deceptive or exploitative circumstances.
In the popular Mission-time Linear Temporal Logic (MLTL) \cite{LVR22}, we can express this as:
$$\texttt{recruited} \rightarrow F_{[0,t]} \texttt{began\_work}$$
where $\texttt{recruited}$ is a Boolean variable that is true if the employee was recruited, $\texttt{began\_work}$ is a Boolean variable that is true if the employee began work, and $F_{[0,t]}$ is the ``eventually'' temporal operator that indicates $\texttt{began\_work}$ should eventually hold within $t$ time units.
Here, the parameter $t$ can be determined with domain specific knowledge, or it
can even be inferred from the data itself.
Our promising initial results with pure Boolean formulas and the scalability of our techniques to modal logics lead us to speculate that this will be a fruitful area to explore.

\subsection{Discussion}
We envision that our work will contribute to wider adoption of neurosymbolic methods in the supply chain domain, and also in other domains where illicit activities are present.
The ultimate goal is to convert the knowledge we gain from this work into actionable change. 
Ideally, this can include informing legislation, helping companies reduce the overhead cost of compliance, guiding law enforcement efforts to identify when to investigate forced labor, and reducing the overall prevalence of forced labor in supply chains.

\begin{acks}
This work is supported by NSF:CCRI Award \#2016592, NSF GRFP Award \#2024364991, and Iowa State Bridging the Divide. 
The authors thank Isaac Blandin and Kendyll Stevenson for their help in data-gathering.
\end{acks}

\bibliographystyle{ACM-Reference-Format}
\bibliography{TACPS25}


\end{document}